\documentclass[journal, twoside]{IEEEtran} 
\usepackage{graphics}
\usepackage{graphicx}
\usepackage{epsfig}
\usepackage{mathptmx}
\usepackage{times}
\usepackage{amsmath,amssymb,amsfonts}
\usepackage{url}
\usepackage{algorithm,algorithmicx}
\usepackage{algpseudocode}
\usepackage{lineno}
\usepackage{hyperref}
\usepackage{bbm}
\newcommand{\norm}[1]{\left\lVert#1\right\rVert}
\usepackage{cite}

\begin{document}
\title{Decision Automation for Electric Power Network Recovery}
\author{Yugandhar~Sarkale,
        Saeed~Nozhati,
        Edwin~K.~P.~Chong,~\IEEEmembership{Fellow,~IEEE,}
        and~Bruce~R.~Ellingwood%
\thanks{This work was supported by the National Science Foundation under Grant CMMI-1638284. This support is gratefully acknowledged. Any opinions, findings, conclusions, or recommendations presented in this material are solely those of the authors and do not necessarily reflect the views of the National Science Foundation. (\textit{Corresponding author: Yugandhar Sarkale}.)}%
\thanks{Yugandhar Sarkale and Edwin K. P. Chong are with Department of Electrical and Computer Engineering, Colorado State University, Fort Collins, CO 80523-1373, USA e-mail: (yugandhar.sarkale@colostate.edu; edwin.chong@colostate.edu).}%
\thanks{Saeed Nozhati is with The B. John Garrick Institute for the Risk Sciences, University of California, Los Angeles, CA 90095, USA e-mail: (snozhati@g.ucla.edu).}%
\thanks{Bruce R. Ellingwood is with Department of Civil and Environmental Engineering, Colorado State University, Fort Collins, CO 80523-1372, USA e-mail: (bruce.ellingwood@colostate.edu).}}


\maketitle

\begin{abstract}
Critical infrastructure systems such as electric power networks, water networks, and transportation systems play a major role in the welfare of any community. In the aftermath of disasters, their recovery is of paramount importance; orderly and efficient recovery involves the assignment of limited resources (a combination of human repair workers and machines) to repair damaged infrastructure components. The decision maker must also deal with uncertainty in the outcome of the resource-allocation actions during recovery. The manual assignment of resources seldom is optimal despite the expertise of the decision maker because of the large number of choices and uncertainties in consequences of sequential decisions. This combinatorial assignment problem under uncertainty is known to be \mbox{NP-hard}. We propose a novel decision technique that addresses the massive number of decision choices for large-scale real-world problems; in addition, our method also features an experiential learning component that adaptively determines the utilization of the computational resources based on the performance of a small number of choices. Our framework is closed-loop, and naturally incorporates all the attractive features of such a decision-making system. In contrast to myopic approaches, which do not account for the future effects of the current choices, our methodology has an anticipatory learning component that effectively incorporates \emph{lookahead} into the solutions. To this end, we leverage the theory of regression analysis, Markov decision processes (MDPs), multi-armed bandits, and stochastic models of community damage from natural disasters to develop a method for near-optimal recovery of communities. Our method contributes to the general problem of MDPs with massive action spaces with application to recovery of communities affected by hazards.
\end{abstract}

\renewcommand*\abstractname{Note to Practitioners}

\begin{abstract}
A significant amount of work has already been done to make communities more resilient against natural or anthropogenic hazards, which can include earthquakes, hurricanes, or a nuclear attack. However, the problem of recovery post-hazard has not been addressed, except in the case of small problems, because of the massive amount of computational resources required to solve the problem. To address this challenging problem, in this work we develop novel decision-making-under-uncertainty algorithms that should be of interest to practitioners in several fields---those that deal with real-world large-scale problem of selecting a single choice given a massive number of alternatives. Even though the applicability of our approach is not limited to the application of community recovery post-hazard, the techniques developed in this work is motivated by, and primarily addresses, post-hazard recovery planning and decision making. We use real-world models of an earthquake to simulate a shock to the community of Gilroy, California. Our automated decision making and optimization framework can be adapted to various types of disasters and different kinds of communities. Our framework accommodates sophisticated stochastic community damage models, handles the stochastic nature of outcomes of human or machine repair actions, takes into account the future impact of current actions, does not suffer from decision fatigue, and incorporates the current policies of the decision makers to automatically plan near-optimal recovery decisions. The current work focuses on optimizing a single recovery policy. In future research, we will extend our framework to simultaneously optimize multiple recovery policies. To further validate the efficacy of our algorithm in dealing with massive stochastic sequential decision-making problems in different domains, future work will include testing the performance of our framework on commercial recommender systems like those used in YouTube and Amazon and large industrial control systems.
\end{abstract}


 \ifCLASSOPTIONpeerreview
 \begin{center} \bfseries EDICS Category: 3-BBND \end{center}
 \fi
%
\IEEEpeerreviewmaketitle

\section{Introduction}\label{Intro}
\IEEEPARstart{A}{utomatic} control systems have had a wide impact in multiple fields, including finance, robotics, manufacturing, and automobiles. Decision automation has gained relatively little attention, especially when compared to decision support systems where the primary aim is to aid humans in the decision-making process. In practice, decision automation systems often do not eliminate human decision makers entirely but rather optimize decision making in specific instances where the automation system can surpass human performance. In fact, human decision makers play a very important role in the selection of models, determining the set of rules, and developing methods that automate the decisions. Nonetheless, decision automation systems remain indispensable in applications where humans are unable to make rational decisions, whether because of the sheer complexity of the system, the enormity of the set of alternatives, or the massive amount of data that must be processed.

Our focus in this paper is to develop a framework that automates decisions for post-disaster recovery of communities. Designing such a framework is ambitious given that it should ideally possess several key properties such as the ability to incorporate sources of uncertainty in the models, information gained at periodic intervals during the recovery process, current policies of the decision-maker, and multiple decision objectives under resource constraints \cite{ress2}. Our framework possesses these desired properties; in addition, our framework uses reasonable computational resources even for massive problems, has the \emph{lookahead} property, and does not suffer from \emph{decision fatigue}.

Civil infrastructure systems, including building infrastructure, power, transportation, and water networks, play a major role in the welfare of any community. The interdependence between the recovery of these networks post-hazard and community welfare addressing the issue of food-security, has been studied in \cite{iEMSs,icasp,saeedinfra}. In this study, we focus on electric power networks (EPNs) because almost all other infrastructure systems rely heavily on the availability of this network. In this study, a stochastic model characterizes the damage to the components of the EPN after an earthquake; similarly, the repair times associated with the repair actions are also given by a stochastic model.

The assignment of limited resources, including repair crews composed of humans and machines, to the damaged components of the EPN after a hazard can be posed as the generalized assignment problem (as defined in \cite{nphard}), which is known to be \mbox{NP-hard}. Several heuristic methods have been demonstrated in the literature to address this problem\cite{heuristic}.

\textbf{Our Contribution:}
Instead of these classical methods, we employ Markov decision processes (MDPs) for the representation and solution of our stochastic decision-making problem, which naturally extends its appealing properties to our framework. In our framework, the solution to the assignment problem formulated as a MDP is computed in an \emph{online} fashion using an approximate dynamic programming method known as \emph{rollout}\cite{online,rollout}. This approach addresses the \emph{curse of dimensionality} associated with large state spaces\cite{adp}. Furthermore, in our framework, the massive action space is handled by using a linear belief model, where a small number of candidate actions are used to estimate the parameters in the model based on a least-squares solution. Our method also employs adaptive sampling inspired by solutions to multi-armed bandit problems to carefully expend the limited simulation budget---a limit on the simulation budget is often a constraint while dealing with large real-world problems. Our approach successfully addresses the goal of developing a technique to deal with problems when the state and actions spaces of the MDP are jointly exceptionally large.

\section{THE ASSIGNMENT PROBLEM}
\subsection{Problem Setup: The Gilroy Community}
The description in this section comes mainly from \cite{saeed}; we give a complete description here for the sake of being self-contained. We describe the EPN of Gilroy, California, which provides the context for our assignment problem. We also briefly discuss the earthquake model, the EPN restoration model, and the computational challenges associated with the assignment problem.
\subsubsection{Network Characterization}\label{test1}
\begin{figure}[t]
	\centering
	\includegraphics[width=\columnwidth]{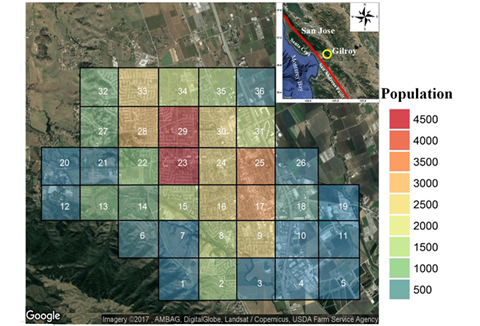}
	\caption{Map of Gilroy's population over the defined grid}
	\label{fig1}
\end{figure}

Gilroy is a moderately-sized growing city located approximately 50 km south of the city of San Jose with a population of 48,821 at the time of the 2010 census \cite{Gilroy1}. The study area is divided into 36 gridded rectangles to define the community and encompasses 41.9 km\textsuperscript{2} area of Gilroy with a population of 47,905. The average number of people per household in Gilroy in 2010 was 3.4, greater than the state and county averages\cite{harnish}. A heat map of the population in the grid is shown in Fig.~\ref{fig1}\cite{ress1}. This model has a resolution that is sufficient to study the methodology at the community level under hazard events. The community is susceptible to severe earthquakes on the San Andreas Fault (SAF).

\begin{figure}[t]
	\centering
	\includegraphics[width=\columnwidth]{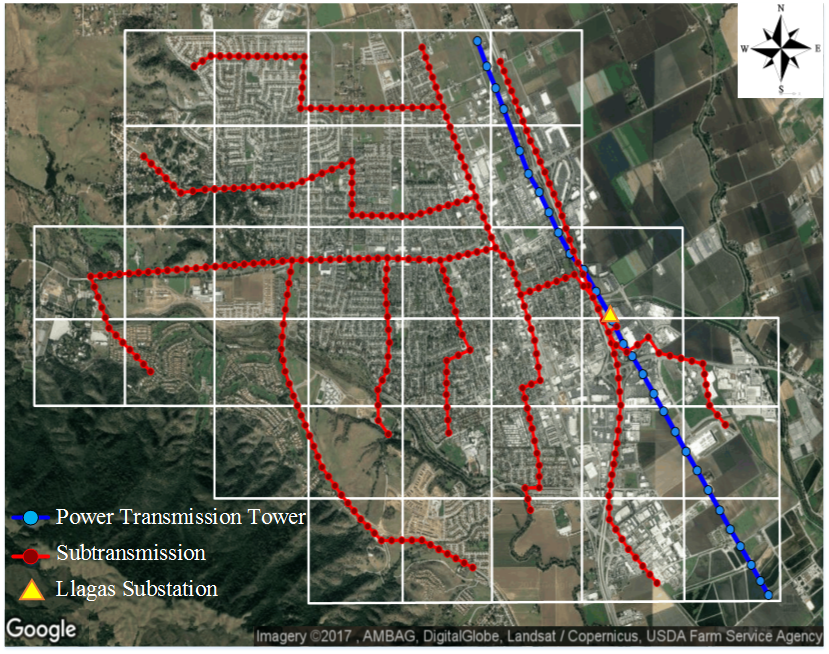}
	\caption{The modeled electric power network of Gilroy}
	\label{fig2}
\end{figure}

The modeled EPN of Gilroy within the defined boundary is shown in Fig.~\ref{fig2}. A 115 kV transmission line supplies the Llagas power substation, which provides electricity to the distribution system. The distribution line components are placed at intervals of 100~m and modeled from the power substation to the centers of the urban grid rectangles. If a component of the EPN is damaged, then along with the damaged EPN component, all the EPN components dependent on the damaged component are rendered nonfunctional or unavailable. If at least one EPN component serving a particular gridded rectangle is unavailable, the entire population of the gridded rectangle does not have electricity.

\subsubsection{Seismic Hazard Simulation}\label{test2}
In this study, we assume that a seismic event of moment magnitude $Mw=6.9$ occurs at the closest points on the SAF projection to downtown Gilroy with an epicentral distance of approximately 12 km\cite{saeed}; this event is similar to the devastating Loma Prieta earthquake of 1989 near Gilroy \cite{Loma}. Ground motion prediction equations (GMPEs) determine the conditional probability of exceeding the ground motion intensity at specific geographic locations within Gilroy given a fault rupture mechanism and epicentral distance for the earthquake \cite{saeedinfra}. We use the Abrahamson et al. \cite{abrahamson} GMPE to estimate the intensity measures (peak ground acceleration) throughout Gilroy.

\subsubsection{Fragility and Restoration Assessment of EPN}\label{test3}
Based on the ground-motion intensities using the above seismic model, we use seismic fragility curves presented in HAZUS-MH\cite{hazus} to calculate the damages to the components of the EPN.
Repair crews, replacement components, and equipment are considered as available units of resources to restore the damaged components of the EPN following the hazard. One unit of resource (RU) is required to repair each damaged component \cite{ouyang}. To restore the EPN, we use the restoration times based on exponential distributions synthesized from HAZUS-MH, as summarized by expected repair times in Table~\ref{T1}.

\begin{table}[h]
	\caption{Expected repair times (Unit: days)}\label{T1}
	\resizebox{\linewidth}{!}{
		\begin{tabular}{llllll}
			\hline
			&Damage States\\
			\hline
			Component & Undamaged & Minor & Moderate & Extensive & Complete \\
			\hline
			Electric sub-station &0 & 1 & 3 & 7 & 30 \\
			Transmission line component &0& 0.5 & 1 & 1 & 2 \\
			Distribution line component &0& 0.5 & 1 & 1 & 1  \\
			\hline
		\end{tabular}
	}
\end{table}


\subsubsection{Challenges}\label{chal}
The total number of modeled EPN components is equal to 327, denoted by $L$. On average, about 60\% of these components are damaged after the simulated earthquake event. At each decision epoch $t=0,1,2,\ldots\,$, the decision maker has to select the assignment of RUs to the damaged components; each component cannot be assigned more than one RU. Note that the symbol $t$ is used to denote a discrete-index representing decision-epoch and is not to be confused with the actual time for recovery. Let the total number of damaged components at any $t$ be represented by $M_t$, and let the total number of RUs be equal to $N$, where $N \ll M_t$ (typically, the number of resource units for repair is significantly less than the damaged components). Then, the total number of possible choices for the assignment at any $t$ is $M_t \choose N$. For 196 damaged components and 29 RUs (15\% of the damaged components), the possible choices at the first decision epoch is approximately $10^{34}$. In addition, the reassignment of all RUs is done when one component gets repaired so that the total number of choices at the second decision epoch is ${195 \choose 29} \approx 10^{34}$.

  Note that the repair time associated with a damaged component will depend on the level of damage, as determined from the fragility analysis described in Section~\ref{test3}. This repair time is random and is exponentially distributed with expected repair times shown in Table~\ref{T1}. Therefore, the outcomes of the repair actions are also random. It is difficult for a human decision maker to anticipate the outcome of repair actions when the outcomes are uncertain; therefore, planning with foresight is difficult. In fact, the problem is difficult to such an extent that assignment of RUs at the first decision epoch itself is challenging. Further, an additional layer of complexity to the problem is manifested owing to the level of damage at each location specified by a probabilistic model\cite{hazus}.

Because of the extraordinarily large number of choices, stochastic initial conditions, and the stochastic behavior of the outcome of the repair actions, our problem has a distinct flavor compared to the generalized assignment problem, and the classical heuristic solutions are not well-suited to this problem. In addition to dealing with these issues, the decision maker has to incorporate the dynamics and the sequential nature of decision making during recovery; thus, our problem represents a stochastic sequential decision-making problem. Last, we would also like our solution to admit most of the desirable properties previously discussed in the Section~\ref{Intro}. Our framework addresses \emph{all} these issues.

\subsection{Problem Formulation}
In this section, we briefly discuss MDPs and the simulation-based representation pertaining to our problem, previously described in \cite{case}, and repeated here for the sake of continuity and completeness. We then specify the components of the MDP for our problem.
\subsubsection{MDP Framework and Simulation-Based Representation}
An MDP is a controlled stochastic dynamical process, widely used to solve disparate decision-making problems. In the simplest form, it can be represented by the 4-tuple $\langle S,A,T,R \rangle$. Here, $S$ represents the set of \emph{states}, and $A$ represents the set of \emph{actions}. The state makes a transition to a new state at each decision epoch (represented by discrete-index $t$) as a result of taking an action. Let $s,s' \in S$ and $a \in A$; then $T$ is the state transition function, where $T(s,a,s')=P(s'\mid s,a)$ is the probability of transitioning to state $s'$ after taking action $a$ in state $s$, and $R$ is the reward function, where $R(s,a,s')$ is the reward received after transitioning from $s$ to $s'$ as a result of action $a$. In our problem, $|S|$ and $|A|$ are finite; $R$ is real-valued and a stochastic function of $s$ and $a$ (deterministic function of $s$, $a$, and $s'$). Implicit in our presentation are also the following assumptions \cite{Puterman}: First-order Markovian dynamics (history independence), stationary dynamics (transition function is not a function of absolute time), and full observability of the state space (outcome of an action in a state might be random, but the state reached is known after the action is completed). The last assumption simplifies our presentation in that we do not need to take actions specifically to reinforce or modify our belief about the underlying state. We assume that recovery actions (decisions) can be taken indefinitely as needed, e.g., until all the damaged components are repaired (infinite-horizon planning). In this setting, we define a \emph{stationary policy} as a mapping $\pi: S \rightarrow A$. Our objective is to find an optimal policy $\pi^*$. For the infinite-horizon case, $\pi^*$ is defined as
\begin{equation}\label{opt}
  \pi^*=\arg\max_{\pi} V^{\pi}(s_0),
\end{equation}
where
\begin{equation}\label{val}
 V^\pi(s_0)=E\left\lbrack\sum_{t=0}^{\infty}\gamma^{\,t}R(s_t,\pi(s_t),s_{t+1})\middle| s_0\right\rbrack
\end{equation}
is called the \emph{value function} for a fixed policy $\pi$, and $\gamma \in (0,1]$ is the discount factor. Note that in~\eqref{opt} we maximize over policies $\pi$, where at each decision epoch $t$ the action taken is $a_t=\pi(s_t)$. Stationary optimal policies are guaranteed to exist for the discounted infinite-horizon optimization criterion \cite{howard}. To summarize, our framework is built on discounted infinite-horizon discrete-time MDPs with finite state and action spaces, though the role $\gamma$ is somewhat tangential in our application.

We now briefly explain the simulation-based representation of an MDP \cite{Fern}. Such a representation serves well for large state, action, and outcome spaces, which is a characteristic feature of many real-world problems; it is infeasible to represent $T$ and $R$ in a simple matrix form for such problems. A simulation-based representation of an MDP is a 4-tuple $\langle S,A,\tilde R,\tilde T \rangle$, where $S$ and $A$ are as before. Here, $\tilde R$ is a stochastic real-valued function that stochastically returns a reward when input $s$ and $a$ are provided, where $a$ is the action applied in state $s$; $\tilde T$ is a \emph{simulator}, which stochastically returns a state sample $s'$ when state $s$ and action $a$ are provided as inputs. We can think of $\tilde R$ and $\tilde T$ as callable library functions that can be implemented in any programming language.

\subsubsection{MDP Specification for EPN Recovery Problem}\label{probform}
\hfill\\
  \textbf{States:} Let $s_t$ denote the state of our MDP at discrete decision epoch $t$: $s_t=(s_t^1,\ldots,s_t^{L}, \rho_t^{1},\ldots,\rho_t^{L})$, $s_t^l$ is the damage state of the $l$th damaged EPN component (the possible damage states are Undamaged, Minor, Moderate, Extensive, and Complete, as shown in Table~\ref{T1}); and $\rho_t^{l}$ is the remaining repair time associated with the $l$th damaged component, where $l \in \{1,\ldots,L\}$. The state transition, and consequently the calculation of $\rho_t^{l}$ and $s_t^{l}$ at each $t$, is explained in the description of simulator $\tilde T$ below.\\
  \textbf{Actions:} Let $a_t$ denote the repair action to be carried out at decision epoch $t$: $a_t=(a_t^1, \ldots, a_t^{L})$, and $a_t^l \in \{0,1\}~\forall l,t$. When $a_t^l=0$, no repair work is to be carried out at $l$th component. Conversely, when $a_t^l=1$, repair work is carried out at the $l$th component. Note that $\sum_{l}a_t^l = N$, and $a_t^l=0$ for all $l$ where $s_t^l$ is equal to Undamaged. Let $D_t$ be the set of all damaged components before a repair action $a_t$ is performed. Let $\mathcal{P}(D_t)$ be the powerset of $D_t$. The total number of possible choices at any decision epoch $t$ is given by $|\mathcal{P}_N(D_t)|$, where
\begin{equation}
  \mathcal{P}_N(D_t)=\{C \in \mathcal{P}(D_t): |C|=N\},
\end{equation}
$|D_t|=M_t$, and $|\mathcal{P}_N(D_t)|= {M_t \choose N}$.\\
  \textbf{Initial State:} The stochastic damage model, previously described in Sections~\ref{test2} and \ref{test3}, is used to calculate the initial damage state $s_0^{l}$. Once the initial damage states of the EPN components are known, depending on the type of the damaged EPN component, the repair times $\rho_0^{l}$ associated with the damaged components are calculated using the mean restoration times provided in Table~\ref{T1}.\\
  \textbf{Simulator $\tilde T$:} Given $s_t$ and $a_t$, $\tilde T$ gives us the new (stochastic) state $s_{t+1}$. We define a \emph{repair completion} as the instant when at least one of the locations where repair work is carried out is fully repaired. The decision epochs occur at these repair-completion times. A damaged component is fully repaired when the damage state of the component changes from any of the four damage states (except the Undamaged state) in Table~\ref{T1} to the Undamaged state. Let us denote the \emph{inter-completion} time by $r_t$, which is the time duration between decision epochs $t$ and $t+1$, and let $\Delta_t=\{\rho_t^l : l\in\{1,\ldots,L\},\ \rho_t^l>0\}$. Then, $r_t=\min\Delta_t$ and $\rho_{t+1}^l = \max(\rho_t^l - r_t, 0)$. Note that it is possible in principle for the repair work at two or more locations to be completed simultaneously, though this virtually never happens in simulation or in practice. When a damaged component is in any of the Minor, Moderate, Extensive, or Complete states, it can only transition directly to the Undamaged state. Instead of modeling the effect of repair via inter-transitions among damage states, the same effect is captured by the remaining repair time $\rho_t$.

      Once a damaged component is restored to the Undamaged state, the RUs previously assigned to it become available for reassignment to other damaged components. Moreover, the RUs at remaining locations, where repair work is unfinished, are also available for reassignment---the repair of a component is \emph{preemptive}. It is also possible for a RU to remain at its previously assigned unrepaired location if we choose so. Because of this reason, preemption of repair work during reassignment is not a restrictive assumption; on the contrary, it allows greater flexibility to the decision maker for planning. Preemptive assignment is known to be particularly useful when an infrastructure system is managed by a central authority, an example of which is EPN \cite{ress2}.

      Even if the same assignment is applied repeatedly to the same system state (let us call this the \emph{current} system state), the system state at the subsequent decision epoch could be different because different components might be restored in the current system state, because of random repair times; i.e., our simulator $\tilde T$ is stochastic. When $M_t$ eventually becomes less than or equal to $N$ because of the sequential application of the repair actions (say at decision epoch $t_a$), the extra RUs are retired so that we have $M_t=N~\forall {t\geq t_{a+1}}$, and the assignment problem is trivial. The evolution of the state of the community as a result of the nontrivial assignments is therefore given by $(s_0,\ldots,s_{t_a})$.\\
  \textbf{Rewards:} We define two \emph{reward functions} corresponding to two different objectives:

  In the first objective, the goal is to minimize the days required to restore electricity to a certain fraction ($\zeta$) of the total population ($p$); recall that for our region of study in Gilroy, $p=47905$. We capture this objective by defining the corresponding reward function as follows:
  \begin{equation}\label{rew1}
    R_1(s_t,a_t,s_{t+1})= r_t,
  \end{equation}
  where we recall that $r_t$ is the inter-completion time between the decision epochs $t$ and $t+1$.
  Let $\hat t_{c}$ denote the decision epoch at which the outcome of repair action $a_{\hat t_{c}-1}$ results in the restoration of electricity to $\zeta \cdot p$ number of people. The corresponding state reached resulting from action $a_{\hat t_{c}-1}$ is $s_{\hat t_{c}}$, called the \emph{goal state} for the first objective. 

  In the second objective, the goal is to maximize the sum (over all the discrete decision epochs $t$) of the product of the total number of people with electricity ($n_t$) after the completion of a repair action $a_t$ and the \emph{per-action time}, defined as the time required ($r_t$) to complete the repair action $a_t$, divided by the total number of days ($t_{\text{tot}}$) required to restore electricity to $p$ people. We capture this objective by defining our second reward function as:
       \begin{equation}\label{rew2}
         R_2(s_t,a_t,s_{t+1})=\frac{n_t \cdot r_t}{t_{\text{tot}}}.
       \end{equation}
        The terms in \eqref{rew2} have been carefully selected so that the product of the terms $n_t$ and $r_t/t_{\text{tot}}$ captures the \emph{impact} of automating a repair action at each decision epoch $t$, in the spirit of maximizing electricity benefit in a minimum amount of time. Let $\tilde t_{c}$ denote the decision epoch at which the outcome of repair action $a_{\tilde t_{c}-1}$ results in the restoration of electricity to the entire population. Then the corresponding goal state is $s_{\tilde t_{c}}$. 

  Note that both $\hat t_{c}$ and $\tilde t_{c}$ need not belong to the set $\{0,\ldots, t_{a-1}\}$, i.e., both $s_{\hat t_{c}}$ and $s_{\tilde t_{c}}$ need not be reached only with a nontrivial assignment. Also, note that our reward function is stochastic because the outcome of each action is random.\\
  \textbf{Discount factor $\gamma$:} A natural consequence of sequential decision making is the problem of \emph{intertemporal choice} \cite{intertemporal}. The problem consists in balancing the rewards and costs at different decision epochs so that the uncertainty in the future choices can be accounted for. To deal with the problem, the MDP model, specifically for our formulation, accommodates a discounted utility, which has been the preferred method of tackling this topic for over a century. In this study, the discount factor $\gamma$ is fixed at 0.99. We have selected a value closer to one because of the use of sophisticated stochastic models described in Sections~\ref{test2} and \ref{test3}; the uncertainty in the outcome of the future choices is modeled precisely via these models, and therefore we can evaluate the value of the decisions several decision-epochs in the future accurately to estimate the impact of the current decision. In our framework, it is possible to select a value closer to zero if the decision automation problem demands the use of simpler models. Moreover, the discounting can be done based on $r_t$---the \emph{real} time required for repair in days (the inter-epoch time)---rather than the number of decision epochs, but this distinction is practically inconsequential for our purposes because of our choice of $\gamma$ being very close to one.

  Next we highlight the salient features of our MDP framework; in particular, we discuss the successful mitigation of the challenges previously discussed in Section~\ref{chal}

Recall that we have a probability distribution for the initial damage state of the EPN components for a simulated earthquake. We generate multiple samples from this distribution to initialize $s_0$ and optimize the repair actions for each of the initial states separately. The outcomes of the optimized repair action for each initial state constitutes a distinct stochastic unfolding of recovery events (recovery path or recovery trajectory). We average over these recovery paths to evaluate the performance of our methods. In our framework, as long as sufficient samples (with respect to some measure of dispersion) are generated, we can appropriately deal with the probabilistic damage-state model.

   Our sequential decision-making formulation also includes modeling the uncertainty in the outcome of repair actions. Thus, our framework can handle both stochastic initial conditions and stochastic repair actions.

   We have formulated the impact of the current decisions on the future choices with exponential discounting. In addition, our sequential decision-making framework addresses the issue of making restoration decisions in stages, where feedback (information) gathered at each stage can play an important role in successive decision making. This is essentially a closed-loop design to compute decisions at each decision epoch.

   Finally, we have defined the second reward function to account for multiple objectives (benefit of electricity ($n_t$) and per-action repair time ($r_t/t_{\text{tot}}$)) without relaxing the constraint on the number of resources.

In the next section, we address the computational difficulties associated with solving the problem, show how to account for the current preferences and policies of the decision maker, and discuss the lookahead property. 

\section{PROBLEM SOLUTION}
\subsection{MDP Solution: Exact Methods}\label{sol}
A solution to an MDP is an optimal policy $\pi^*$. There are several methods to exactly compute $\pi^*$; here, we discuss the \emph{policy iteration} algorithm because it bears some relationship with the \emph{rollout} method, which we describe later.

Suppose that we have access to a nonoptimal policy $\pi$. The value function for this policy $\pi$ in \eqref{val} can be written as
\begin{equation}\label{bellu}
  V^\pi(s)= R(s,\pi(s))+\gamma \sum_{s'} P(s'\mid s,\pi(s))\cdot V^\pi(s')~\forall s \in S,
\end{equation}
where $V^\pi$ can be calculated iteratively using the \emph{Bellman's update equation} or by solving a linear program \cite{belllin}. This calculation of $V^\pi$ is known as the policy \emph{evaluation} step of the policy iteration algorithm.
The $Q$ value function of policy $\pi$ is given by
\begin{equation}\label{Qval}
  Q_{\pi}(s,a)=R(s,a)+\gamma \sum_{s'} P(s'\mid s,a)\cdot V^\pi(s'),
\end{equation}
which is the expected discounted reward in the future after starting in some state $s$, taking action $a$, and following policy $\pi$ thereafter.
An improved policy $\pi'$ can be calculated as
\begin{equation}\label{roll}
  \pi'(s_t)=\arg\max_{a_t}Q_\pi(s_t,a_t).
\end{equation}
The calculation of an improved policy in \eqref{roll} is known as the policy \emph{improvement} step of the policy iteration algorithm. Even if the policy $\pi'$ defined in \eqref{roll} is nonoptimal, it is a \emph{strict} improvement over $\pi$ \cite{howard}. This result is called the \emph{policy improvement theorem}. Note that the improved policy $\pi'$ is generated by solving, at each state $s$, an optimization problem with $Q_\pi(s,\cdot)$ as the objective function. In the policy iteration algorithm, to compute the optimal policy $\pi^*$, the policy evaluation and improvement steps are repeated iteratively until the policy improvement step does not yield a strict improvement.

Unfortunately, algorithms to compute the exact optimal policy are intractable for even moderate-sized state and actions spaces. Each iteration of the policy evaluation step requires $\mathcal{O}(|S|^3)$ time using a linear program and $\mathcal{O}(|S||A|)$ time using Bellman's update for a given $\pi$.\footnote{If the policy evaluation step is done using the Bellman's update with a given $\pi$, instead of solving a linear program, the algorithm is called a \emph{modified} policy iteration; conventionally, the term \emph{policy iteration} is used only when the policy evaluation step is performed by solving a linear program.} In the previous example from Section~\ref{chal}, where the total number of damaged components after the initial shock is equal to 196, for the five damage states in Table~\ref{T1} and two repair actions (repair and no-repair), $|S|=5^{196}$ and the $|A|=2^{196}$. Note that our state and action space is jointly massive. In our case, and for other large real-world problems, calculating an exact solution is practically impossible; even enumerating and storing these values in a high-end supercomputer equipped with state-of-the-art hardware is impractical.
\subsection{Rollout: Dealing with Massive S}
We now motivate the rollout algorithm \cite{rollout} in relation to our simulation-based framework and the policy iteration algorithm.

When dealing with large $S$ and $A$, approximation techniques have to be employed given the computational intractability of the exact methods. A general framework of using approximation within the policy iteration algorithm is called \emph{approximate policy iteration}---rollout algorithms are classified under this framework \cite{lagoudakis2003reinforcement}. In rollout algorithms, usually the policy evaluation step is performed approximately using Monte Carlo sampling and the policy improvement step is exact. The policy improvement step is typically exact, at some computational cost, because approximating the policy improvement step requires the use of sophisticated techniques tailored to the specific problem being solved by rollout to avoid poor solution quality. A novel feature of our work is that we approximate both the policy improvement and policy evaluation step. The approximation to the policy improvement step is explained in Section~\ref{lin}.

The policy evaluation step is approximated as follows. An implementable (in a programming sense) stochastic function (simulator) $SimQ(s_t,a_t,\pi,h)$ is defined in such a way that its expected value is $Q_\pi(s_t,a_t,h)$, where  $Q_\pi(s_t,a_t,h)$ denotes a finite-horizon approximation of $Q_\pi(s_t,a_t)$, and $h$ is a finite number representing horizon length. In the rollout algorithm, $Q_\pi(s_t,a_t,h)$ is calculated by simulating action $a_t$ in state $s_t$ and thereafter following $\pi$ for another $h-1$ decision epochs, which represents the approximate policy evaluation step. This is done for candidate actions $a_t \in A(s_t)$, where $A(s_t)$ is the set of all the possible actions in the state $s_t$. A finite-horizon approximation $Q_\pi(s_t,a_t,h)$) is unavoidable because, in practice, it is of course impossible to simulate the system under policy $\pi$ for an infinite number of epochs. Recall, however, that $V^\pi(s_t)$, and consequently $Q_\pi(s_t,a_t)$, is defined over the infinite horizon. It is easy to show the following result \cite{Fern}:
\begin{equation}\label{Qapprox}
  \left | Q_\pi(s_t,a_t)- Q_\pi(s_t,a_t,h)\right |=\frac{\gamma^{\,h}R_{\text{max}}}{1-\gamma},
\end{equation}
where $R_{\text{max}}$ is the largest value of the reward function (either $R_1$ or $R_2$).
The approximation error in \eqref{Qapprox} reduces exponentially fast as $h$ grows. Therefore, the $h$-horizon calculation appropriately approximates the infinite-horizon version, for we can always choose $h$ sufficiently large such that the error in \eqref{Qapprox} is arbitrarily small. The algorithm for rollout and the simulator is presented in Algorithms~\ref{rollout} and \ref{sim}, respectively, where $\alpha=|A(s_t)|$, $a_{t,i} \in A(s_t)$ (here $i \in \{1,\ldots,\alpha\}$),  and $\beta$ is the total number of samples available to estimate $Q_\pi(s_t,a_t,h)$. Algorithm~\ref{rollout} is also called a \emph{uniform} rollout algorithm because $\beta$ samples are allocated to each action $a_t$ in $A(s_t)$ uniformly. In essence, rollout uses Monte-Carlo simulations in the policy evaluation step to calculate approximate $Q$ values; the quality of the approximation is often practically good enough even for small $h$.
\begin{algorithm}
    \caption{\textbf{Uniform\_Rollout}($\pi, h, \beta, s_t, A(s_t)$)}
    \label{rollout}
    \begin{algorithmic}
    \For{$i=1$ to $\alpha$}
    \For{$j=1$ to $\beta$}
    \State $ Q^{i,j} \gets \textbf{SimQ}(s_t,a_{t,i},\pi,h)$ \Comment{See algorithm 2}
    \EndFor
     \State  $ Q_t(i) \gets \emph{Average}(Q^{i,j}$) \Comment{With respect to $j$}
    \EndFor
    \State  $k \gets \arg\max_{i} Q_t$
    \State \Return $a_{t,k}$
    \end{algorithmic}
    \end{algorithm}
\begin{algorithm}
    \caption{Simulator \textbf{SimQ}$(s_t, a_{t,i}, \pi, h)$}
    \label{sim}
    \begin{algorithmic}
    \State $t'=0$
    \State $s'_0\gets s_t$
    \State $s'_{t'+1}\gets \tilde T(s'_{t'},a_{t,i})$
    \State $r \gets \tilde R(s'_{t'},a_{t,i},s'_{t'+1})$
    \For{$\lambda=1$ to $h-1$}
    \State $s'_{t'+1+\lambda} \gets \tilde T(s'_{t'+\lambda},\pi(s'_{t'+\lambda}))$
    \State $r\gets r+ \gamma^{\,\lambda}\tilde R(s'_{t'+\lambda},\pi(s'_{t'+\lambda}),s'_{t'+1+\lambda})$
    \EndFor
    \State \Return $r$
    \end{algorithmic}
    \end{algorithm}

 Rollout fits well in the paradigm of online planning. In online planning, the optimal action is calculated only for the current state $s_t$, reducing the computational effort associated with a large state space. Similarly, in our problem, we need to calculate repair actions for the current state of the EPN without wasting computational resources on computing repair actions for the states that are never encountered during the recovery process. Therefore, the property of online planning associated with Algorithm~\ref{rollout} is important for recovery, and even if the policy $\pi$ (called the \emph{base policy} in the context of Algorithm~\ref{rollout}) is applied repeatedly (``rolled out'') for $h-1$ decision epochs, we focus only on the \emph{encountered} states as opposed to dealing with \emph{all} the possible states (cf., \eqref{bellu}). In essence, for the recovery problem, rollout can effectively deal with large sizes of the state space because the calculation of the policy is amortized over time.

Consider the following example. In the context of online planning, for the sake of argument suppose that the action space has only a single action. Even for such a superficially trivial example, the outcome space can be massive. However, the representation of the problem in our framework limits the possible outcomes for any $(s,a)$ pair to $N$, bypassing the problem with the massive outcome space.

We can use existing policies of expert human decision makers as the base policy in the rollout algorithm. The ability of rollout to incorporate such policies is reflected by its interpretation as one-step of policy iteration, which itself starts from a nonoptimal policy $\pi$. In fact, rollout as described here is a ``one-step lookahead" approach 
(here, one-step lookahead means one application of policy improvement) \cite{rollout}. Despite the stochastic nature of the recovery problem, the \emph{uniform} rollout algorithm (as defined by Algorithm~\ref{rollout}) computes the expected future impact of every action to determine the optimized repair action at each $t$. Because the policy evaluation step is approximate, rollout cannot guarantee a \emph{strict} improvement over the base policy; however, the solution obtained using rollout is never worse than that obtained using the base policy \cite{rollout} because we can always choose the value of $h$ and $\beta$ such that the rollout solution is no worse than the base policy solution \cite{Dimitrakakis2008b}. In practice, compared to the \emph{accelerated policy gradient} techniques, rollout requires relatively few simulator calls (Algorithm~\ref{sim}) to compute equally good near-optimal actions \cite{accelerate}.

\subsection{Linear Belief Model: Dealing with Massive A}\label{lin}
The last remaining major bottleneck with the rollout solution proposed above is that for any state $s_t$, to calculate the repair action, we must compute the $argmax$ of the $Q$ function at $s_t$. This involves evaluating the $Q$ values for candidate actions and searching over the space of feasible actions. Because of online planning, we no longer deal with the entire action space $A$ but merely $A(s_t)$. For the example previously discussed in Section~\ref{chal}, even though this is a reduction from $2^{196}$ to $196\choose29$, the required computation after the reduction remains substantial.

Instead of rolling out all $a_t \in A(s_t)$ exhaustively, we train a set of parameters of a linear belief model (explained below) based on a small subset of $A(s_t)$, denoted by $\tilde A(s_t)$. The elements of $\tilde A(s_t)$, denoted by $\tilde a_t$, are chosen randomly, and the size of the set $\tilde A(s_t)$, denoted by $\tilde \alpha$, is determined in accordance with the simulation budget available at each decision epoch $t$. The simulation budget $B$ at each decision epoch will vary according to the computational resources employed and the run-time of Algorithm~\ref{sim}. Thereafter, $a_t$ is calculated using the estimated parameters of the linear belief model.

Linear belief models are popular in several fields, especially in drug discovery \cite{free}. Given an action $\tilde a_{t,i}$ selected from $\tilde A(s_t)$, the linear belief model can be represented as
\begin{equation}\label{linsum}
 \tilde Q^{i,j} = \sum_{n=1}^{N}\sum_{m=1}^{M} \mathbf{X}_{mn}\cdot \Theta_{mn} + \eta_{mn},
\end{equation}
where
\begin{equation}\label{mod}
\mathbf{X}_{mn}=
    \begin{cases}
      1 & \text{if \textit{n}th RU is assigned to \textit{m}th location}\\
      0 & \text{otherwise,}
    \end{cases}
\end{equation}
 $i \in \{1,\ldots,\tilde\alpha\}$, $j \in \{1,\ldots,\beta\}$, $\tilde Q^{i,j}$ are the $Q$ values corresponding to $\tilde a_{t,i}$ obtained with Algorithm~\ref{sim}, and $\eta_{mn}$ represents noise. 
 Let $\tilde Q^i=\frac{1}{\beta}\sum_{j=1}^{\beta}\tilde{Q}^{i,j}$.
  In this formulation, each parameter $\Theta_{mn}$ additively captures the impact on the $Q$ value of assigning a RU (indexed by $n$) to a damaged component (indexed by $m$). In particular, the contribution of each parameter is assumed to be independent of the presence or absence of the other parameters (see the discussion at the end of this section). Typically, linear belief models include an additional parameter: the constant intercept term $\Theta_0$ so that \eqref{linsum} would be expressed as
   \begin{equation}\label{linsum2}
 \tilde Q^{i,j} = \Theta_0 + \sum_{n=1}^{N}\sum_{m=1}^{M} \mathbf{X}_{mn}\cdot \Theta_{mn} + \eta_{mn}.
\end{equation}
   However, our model excludes $\Theta_0$ because it would carry no corresponding physical significance unlike the other parameters.

   The linear belief model in~\eqref{linsum} can be equivalently written as
  \begin{equation}\label{mat}
    \mathbf{y}=\mathbf{H}\cdot\theta+\eta,
  \end{equation}
  where $\mathbf{y}$ (of size $\tilde \alpha \times 1$) is a vector of the $\tilde Q^i$ values calculated for all the actions $\tilde a_t \in \tilde A(s_t)$, $\mathbf{H}$ (of size $\tilde \alpha \times (M_t \cdot N)$) is a binary matrix where the entries are in accordance with \eqref{linsum}, \eqref{mod}, and the choice of set $\tilde A(s_t)$, $\theta$ (of size $(M_t \cdot N) \times 1$) is a vector of parameters $\Theta_{mn}$, and $\eta$ (of size $(M_t \cdot N) \times1$) is the noise vector. The simulation budget $B$ at each decision epoch is divided among $\tilde \alpha$ and $\beta$ such that ${B=\tilde\alpha\cdot\beta}$. In essence, based on the $\tilde a_t \in \tilde A(s_t)$---which corresponds to the assignment of $N$ RUs to $M_t$ damaged components according to~\eqref{mod}---the matrix $\mathbf{H}$ is constructed. The vector $\mathbf{y}$ is constructed by computing the $Q$ values corresponding to $\tilde a_t$ according to Algorithm~\ref{sim}.

  We estimate the parameter vector $\hat{\theta}$ by solving the least squares problem of minimizing $\|\mathbf{y}-\mathbf{H}\hat{\theta}\|_2$ with respect to $\hat{\theta}$. We chose a least squares solution to estimate $\hat{\theta}$ because least-squares solutions are well-established numerical solution methods, and if the noise is an uncorrelated Gaussian error, then $\hat{\theta}$ estimated by minimizing $\|\mathbf{y}-\mathbf{H}\hat{\theta}\|_2$ is the \emph{maximum likelihood estimate}. In our framework, the rank of $\mathbf{H}$ is $(M_t\cdot N)-(N-1)$. Therefore, the estimated parameter vector $\hat \theta$, which consists of parameters $\hat \Theta_{mn}$ and is calculated using the ordinary least squares solution, is not unique and admits an infinite number of solutions \cite{kailath}. Even though $\hat{\theta}$ is not unique, $\mathbf{\hat y}$ defined by the equation $\mathbf{\hat y}=\mathbf{H}\cdot \hat \theta$ is unique; moreover, the value of $\norm{\mathbf{y}-\mathbf{H}\cdot \hat \theta}_2^2$ is unique. We can solve our least squares problem uniquely using either the Moore-Penrose pseudo-inverse or singular value decomposition by calculating the minimum-norm solution \cite{chong}. In this work, we have used the Moore-Penrose pseudo-inverse. Note that ${\tilde{\alpha}\gg (M_t\cdot N)-(N-1)}$ (the number of rows of the matrix $\mathbf{H}$ is much greater than its rank).

  Once the parameters $\hat{\Theta}_{mn}$ are estimated, the optimum assignment of the RUs is calculated successively (one RU at a time) depending on the objective in \eqref{rew1} and \eqref{rew2}. In the calculation of the successive optimum assignments of RU in Algorithm~\ref{rolloutwbelief}, let $\hat m$ denote the assigned location at each RU assignment step; then all the estimated parameters corresponding to $\hat m$ (denoted by parameters $\hat \Theta_{\hat m,index}$, where $index \in \{1,\ldots,N\}$) are set to $\infty$ or $-\infty$ depending on \eqref{rew1} and \eqref{rew2}, respectively. This step ensures that only a single RU is assigned at each location. This computation is summarized in Algorithm~\ref{rolloutwbelief}. Similar to Algorithm~\ref{rollout}, the assignment of $\beta$ samples to every action in $\tilde A(s_t)$ is uniform.
  \begin{algorithm}
    \caption{\textbf{Uniform\_Rollout w/ Linear\_Belief} ($\pi, h, \beta, s_t, \mathbf{H}, \tilde A(s_t)$)}
    \label{rolloutwbelief}
    \begin{algorithmic}
    \State \textbf{Intialize} $a_t=[\mathbf{0}]$
    \For{$i=1$ to $\tilde \alpha$}
    \For{$j=1$ to $\beta$}
    \State $ \tilde Q^{i,j} \gets \textbf{SimQ}(s_t,\tilde a_{t,i},\pi,h)$ \Comment{See algorithm 2}
    \EndFor
    \State $ y(i) \gets \emph{Average}(\tilde Q^{i,j}$) \Comment{With respect to $j$}
    \EndFor
    \State $\hat \theta\gets\textbf{OLS}( y,\mathbf{H})$ \Comment{Ordinary least squares solution}
    \For{$k=1$ to $N$} \Comment{RU assignment step begins}
    \State $ (\hat m, \hat n)\gets\arg\min_{m,n}\hat\theta $ \Comment{Min for~\eqref{rew1} and max for~\eqref{rew2}}
    \State $a_t^{\hat m}\gets1$
    \For{$index=1$ to $N$}
    \State $\hat \Theta_{\hat m,index} \gets \infty$ \Comment{$-\infty$ for~\eqref{rew2}}
    \EndFor
    \EndFor
    \State \Return $a_t$
    \end{algorithmic}
    \end{algorithm}

Our Algorithm~\ref{rolloutwbelief} has several subtleties, as summarized in the following discussion.

The use of linear approximation for dynamic programming is not novel in its own right (it was first proposed by Bellman et al. \cite{bellmanlin}). The only similarity between the typical related methods (described in \cite{lagoudakis2003reinforcement}) and our approach is that we are fitting a linear function over the rollout values---the belief model is a function approximator for the $Q$ value function in Algorithm~\ref{rollout}---whereas the primary difference is explained next.

  Most of the error and convergence analyses for MDPs use the max-norm ($\mathcal{L}_\infty$ norm) to guarantee performance; in particular, the performance guarantee on the policy improvement step in \eqref{roll} and the computation of $a_t$ using rollout in Algorithm~\ref{rollout} are two examples. It is possible to estimate the parameters $\hat \theta$ to optimize the $\mathcal{L}_\infty$ norm by solving the resultant optimization problem using linear programming (see \cite{Stiefel}). The influence of estimating $\hat \theta$ to optimize the $\mathcal{L}_\infty$ norm, when a linear function approximator is used to approximate the $Q$ value function, on the error performance of any algorithm that falls in the general framework of approximate policy iteration is analyzed in \cite{Guestrin}.\footnote{Instead of formulating the approximation of the $Q$ value function as a regression problem, it is also possible to pose the $Q$  value function approximation as a classification problem.\cite{lagoudakis2003reinforcement}} Our approach is different from such methods because in our setting, the least squares solution optimizes the $\mathcal{L}_2$ norm, which we found to be advantageous.

Indeed, our solution shows promising performance. Three commonly used statistics to validate the use of the linear-belief model and the least squares solution in Algorithm~\ref{rolloutwbelief} are as follows: residual standard error (RSE), R-squared ($R^2$), and F-statistic.
The RSE for our model is $10^{-5}$, which indicates that the linear model satisfactorily fits the $Q$ values computed using rollout.
The $R^2$ value for our model is 0.99, which indicates that the computed features/predictors ($\hat \theta$) can effectively predict the $Q$ values.
The F-statistic is 4 (away from 1) for a large $\tilde \alpha$ ($\tilde \alpha=10^6$; whereas, at each $t$, the rank of $\mathbf{H}$ is never greater than 5850), which indicates that the features/predictors defined in \eqref{linsum} and \eqref{mod} are statistically significant. We can increase the number of predictors by including the interactions between the current predictors at the risk of overfitting the $Q$ values with the linear model \cite{occam}. As the authors in \cite{lagoudakis2003reinforcement} aptly point out, ``increasing expressive power can lead to a surprisingly worse performance, which can make feature engineering a counterintuitive and tedious task."

\subsection{Adaptive Sampling: Utilizing Limited Simulation Budget}
Despite implementing best software practices to code fast simulators and deploying the simulators on modern supercomputers, the simulation budget $B$ is a precious resource, especially for massive real-world problems. A significant amount of research has been done in the simulation-based optimization literature \cite{jia, sg, hx, sg2} to manage simulation budget. The related methods have also been demonstrated on real-world problems \cite{gp, case}.

A classic simulation-based approach such as optimal computing budget allocation \cite{Chen2000} is not employed here to manage budget, instead the techniques in our study are inspired by solutions to the multi-armed bandit problems \cite{bandit,dar,Auer,ucb}, which are topical in the computer science and artificial intelligence community, especially in research related to \emph{reinforcement learning}. The problem of (managing budget) expending limited resources is studied in reinforcement learning, although in a completely different context, where few optimal choices must be selected among a large number of options to optimize a stochastic objective function.

It has been our observation that two independent research communities---simulation-based optimization and computer science---have worked on similar problems in isolation. In this work, our solutions have been inspired by the later approach and will serve to bridge the gap between the work in the two research communities.

Algorithm~\ref{rollout}, and consequently also Algorithm~\ref{rolloutwbelief}, is not only directly dependent upon the speed of Algorithm~\ref{sim} (simulator) but also requires an accurate $Q$ value function estimate to guarantee performance. Therefore, typically a huge sampling budget in the form of large $\beta$ is allocated uniformly to every action $\tilde a_t \in \tilde A(s_t)$. This naive approach decreases the value of  $\tilde \alpha$ (which is the size of the set $\tilde A(s_t)$);\footnote{Note that $B$ is fixed and depends on the simulator runtime and the computational platform on which the algorithm runs. Recall that $B=\tilde \alpha \cdot \beta$, and the larger the value of $\beta$ required to guarantee performance, the smaller the value of $\tilde \alpha$.} consequently, the parameter vector $\theta$ is trained on a smaller number of $Q$ values. In practice, we would like to get a rough estimate of the $Q$ value associated with every action in the set $\tilde A(s_t)$ and adaptively spend the remaining simulation budget in refining the accuracy of the $Q$ values corresponding to the best-performing actions; this is the \emph{exploration vs. exploitation} problem in optimal learning and simulation optimization problems\cite{WSC}. Spending the simulation budget $B$ in a nonuniform, adaptive fashion in the estimation of the $Q$ value function would not only train the parameter vector $\theta$ on a larger size of the set $\tilde A(s_t)$ via the additive model in \eqref{linsum} but also train the parameters $\Theta_{mn}$ on $Q$ values corresponding to superior actions (this is because in an adaptive scheme, $B$ is allocated in refining the accuracy of only those actions that show promising performance), consequently refining the accuracy of the parameters. The nonuniform allocation of simulation budget is the experiential learning component of our method, which further enhances Algorithm~\ref{rolloutwbelief}.

An interesting closed-loop sequential method pertaining to drug discovery that bears some resemblance to the experiential learning component of our method is described in \cite{knowledge}, where the alternatives (actions are called alternatives in their work) are selected adaptively using \emph{knowledge gradient} (KG). Further, in their work, KG is combined with a linear-belief model, and the results are demonstrated on a moderate-sized problem. Unfortunately, the algorithms proposed in \cite{knowledge} are not directly applicable to our problem because the algorithms in \cite{knowledge} necessitate sampling over the actions in $A(s_t)$, instead of $\tilde A(s_t)$.

Instead of uniformly allocating $\beta$ samples to each action in Algorithm~\ref{rollout}, nonuniform allocation methods have been explored in the literature to manage the rollout budget \cite{Dimitrakakis2008b}.  An analysis of performance guarantees for nonuniform allocation of the rollout samples remains an active area of research \cite{dimitri2008a}. However, we extend the ideas in \cite{Dimitrakakis2008b} and \cite{dimitri2008a}, pertaining to nonuniform allocation, to Algorithm~\ref{rolloutwbelief} based on the theory of \emph{multi-armed bandits}.

In bandit problems, the agent has to sequentially allocate resources among a set of bandits, each one having an unknown reward function, so that a \emph{bandit objective} \cite{bandit} is optimized. There is a direct overlap between managing $B$ and the resource allocation problem in multi-armed bandit theory; the allocation of the simulation budget $B^*$ defined by the equation $B^*=B-\tilde\alpha$ sequentially to the state-action pair $(s_t,\tilde a_t)$ during rollout is equivalent to a variant of the classic multi-armed bandit problem \cite{Dimitrakakis2008b}.

In this study, we consider two bandit objectives: probable approximate correctness (PAC)  and cumulative regret. In the \emph{PAC} setting, the goal is to allocate budget $B^*$ sequentially so that we find a near-optimal ($\epsilon$ of optimal) action $\tilde a_t$ with high probability ($1-\delta$) when the budget $B^*$ is exhausted. Algorithm~\ref{rollout} is PAC optimal when $h$ and $\beta$ are selected in accordance with the \emph{fixed algorithm} in \cite{dimitri2008a}. For our decision-automation problem, the value of $\beta$ required to guarantee performance is typically large. Nonuniform allocation algorithms like \emph{median elimination} are PAC optimal \cite{dar} (the median elimination algorithm is asymptotically optimal, so no other nonuniform resource-allocation algorithm can outperform the median elimination algorithm in the worst case). However, the choice of $(\epsilon,\delta)$ for the PAC objective is arbitrary; therefore, the PAC objective is not well-suited to our decision automation problem. Further, the parameters of the median elimination algorithm that guarantee performance are directly dependent on the $(\epsilon,\delta)$ pair.

The second common objective function in bandits problems mentioned earlier, \emph{cumulative regret} is well-suited to our problem. During the optimization of \emph{cumulative regret}, the budget $B^*$ is allocated sequentially in such a way that when the budget is exhausted, the expected total reward is very close to the best possible reward (called minimizing the cumulative regret). An algorithm in \cite{Auer} called \emph{UCB1} minimizes the cumulative regret; in fact, no other algorithm can achieve a better cumulative expected regret (in the sense of scaling law). Usually, cumulative regret is not an appropriate objective function to be considered in nonuniform rollout allocation \cite{ucb} because almost all common applications require finding the best (approximately) action $a_t$, whereas in our problem, we would like to allocate the budget nonuniformly so that the parameter vector $\hat \theta$ in Algorithm~\ref{rolloutwbelief} is estimated in the most efficient way. Therefore, it is natural to allocate the computing budget so that the expected cumulative reward over all the $\tilde a_t$ ($Q$ values in the vector $\mathbf{y}$ in Algorithm~\ref{rolloutwbelief}) is close to the optimal value.

Based on the simulator runtime, the underlying computational platform, and the actual time provided by the decision maker to our automation system, suppose that we fix $B$ and in turn the size of the set $\tilde A(s_t)$. We exhaust a budget of $\tilde \alpha$ samples (one per action) from $B$ on getting rough estimates of the $Q$ value function for the entire set $\tilde{A}(s_t)$; the remaining budget $B-{\tilde\alpha}$ (denoted by $B^*$) is allocated adaptively using the UCB1 algorithm. This scheme of adaptively managing $B^*$ in Algorithm~\ref{rolloutwbelief} is summarized in Algorithm~\ref{rolloutwbeliefa}.

Algorithm~\ref{rolloutwbeliefa} alleviates the shortcomings of Algorithm~\ref{rolloutwbelief} by embedding the experiential learning component using the UCB1 algorithm. The UCB1 algorithm assumes that the rewards lie in the interval [0,1]. Satisfying this condition is trivial in our case because the rewards are bounded and thus can be always normalized so that they lie in the interval [0,1]; it is important to implement the normalization of $\tilde R$ in Algorithm~\ref{sim} when we use Algorithm~\ref{rolloutwbeliefa}. In Algorithm~\ref{rolloutwbeliefa}, not only is $B^*\gg\beta$, but we can also select $\tilde \alpha$ larger than that in Algorithm~\ref{rolloutwbelief} and train the parameter vector $\theta$ on a larger size of the set $\tilde A(s_t)$, which in turn will yield better estimates of $\hat \theta$. Note that Algorithm~\ref{rolloutwbeliefa} does not merely manage the budget $B^*$ adaptively (adaptive rollout), but it also handles massive action spaces through the linear belief model described in Section~\ref{lin} (this is because Algorithm~\ref{rolloutwbeliefa} is Algorithm~\ref{rolloutwbelief} with the UCB1 step appended).

In essence, Algorithm~\ref{rolloutwbeliefa} has three important steps: First, $Q$ values corresponding to $\tilde \alpha$ actions in the set $\tilde A(s_t)$ are computed. Second, the estimates for the $Q$ values corresponding to the most promising actions are refined by nonuniform allocation of the simulation budget using the UCB1 algorithm. Last, based on the ordinary least squares solution to calculate $\hat \theta$, the RUs are assigned sequentially just like in Algorithm~\ref{rolloutwbelief} described in Section~\ref{lin}.
\begin{algorithm}
    \caption{\textbf{Adaptive\_Rollout w/ Linear\_Belief} ($\pi, h, B^*, s_t, \mathbf{H}$)}
    \label{rolloutwbeliefa}
    \begin{algorithmic}
    \State \textbf{Intialize} $a_t=[\mathbf{0}]$
    \For{$i=1$ to $\tilde \alpha$}
    \State $ \tilde y(i) \gets \textbf{SimQ}(s_t,\tilde a_{t,i},\pi,h)$ \Comment{See algorithm 2}
    \EndFor
    \State $Count\gets\tilde \alpha$
    \State $Count_{i}\gets[\mathbf{1}]$ \Comment{Counts the number of samples assigned to the $i$th action}
    \While{$B^*$ is not zero} \Comment{UCB1 step}
    \For{$i=1$ to $\tilde \alpha$}
    \State $ d(i) \gets \tilde y(i)+\sqrt{\frac{2\ln (Count)}{Count_i(i)}}$
    \EndFor
    \State $\tau\gets\arg\max_i d$
    \State $Count_i(\tau)\gets Count_i(\tau)+1$
    \State $Count\gets Count+1$
    \State $ \tilde y(\tau) \gets \frac{(Count_i(\tau)-1)\cdot \tilde y(\tau)+ \textbf{SimQ}(s_t,a_{t,\tau},\pi,h)}{Count_i(\tau)}$
    \State $B^*\gets B^*-1$
    \EndWhile
    \State $\hat \theta=\textbf{OLS}( \tilde y,\mathbf{H})$ \Comment{Ordinary least squares solution}
    \For{$k=1$ to $N$}
    \State $ (\hat m, \hat n)\gets\arg\max_{m,n}\hat\theta $ \Comment{Min for~\eqref{rew1} and max for~\eqref{rew2}}
    \State $a_t^{\hat m}\gets1$
    \For{$index=1$ to $N$}
    \State $\hat \Theta_{\hat m,index} \gets -\infty$ \Comment{$\infty$ for~\eqref{rew1}}
    \EndFor
    \EndFor
    \State \Return $a_t$
    \end{algorithmic}
    \end{algorithm}

\section{SIMULATION RESULTS: MODELING GILROY RECOVERY}\label{simres}
We simulate 25 different damage scenarios (stochastic initial conditions) for each of the figures presented in this section. Calculation of the recovery for a single damage scenario is computationally expensive. Nevertheless, multiple initial conditions are generated to deal with the stochastic earthquake model as discussed in Section~\ref{probform}. In case of both Objective 1 and Objective 2, corresponding to $R_1$ and $R_2$ respectively, there will be a distinct recovery path for each of the initial damage scenarios. To present the results for Objective 1, we do not explicitly show the recovery trajectories. We are only interested in the number of days it takes to provide maximum benefit in the sense of optimizing $R_1$. Therefore, the results are presented in terms of a cumulative moving average plot. In Objective 2, for both Algorithm~\ref{rolloutwbelief} and Algorithm~\ref{rolloutwbeliefa}, the recovery computed using these algorithms outperform the base policy for every single scenario.

There are several candidates for determining the base policy to be used in the simulation. For a detailed discussion on these candidates in post-hazard recovery planning, see \cite{ress1}. For the simulations presented in this study, a random base policy is used. The total number of RUs are capped at 15\% of the damaged components for each scenario. The maximum number of damaged components in any scenario encountered in this study is 205, i.e., the size of the assignment problem at any $t$ is less than $10^{37}$. The simulators have a runtime of $10^{-5}$~s when $h=1$, and this runtime varies with the parameter $h$. The deeper we rollout the base policy in any variation of the rollout algorithm, the larger the simulation time per-action and the smaller the action space covered to train our parameters.

For Algorithm~\ref{rolloutwbelief} and the computational platform (AMD EPYC 7451, 2.3 GHz, and 96 cores), the value of $\beta$ is capped at 100 and the value of $\tilde \alpha$ is capped at $10^6$. Note that it is possible to parallelize Algorithm~\ref{rolloutwbelief} at two levels. The recovery of each damage scenario can be computed on a different processor, and then their average can be calculated. Further, Algorithm~\ref{rolloutwbelief} offers the opportunity to parallelize over $\tilde A(s_t)$ because a uniform budget can be allocated to a separate processor to return the average $Q$ value for each $\tilde a(s_t)$. On the contrary, the allocation of budget $B^*$ in Algorithm~\ref{rolloutwbeliefa} is sequential, and only a single $Q$ value corresponding to the allocated sample is evaluated (see the UCB1 step in Algorithm~\ref{rolloutwbeliefa}). Based on the updated $Q$ value (calculation of $\tilde y(\tau)$ in Algorithm~\ref{rolloutwbeliefa}), further allocation is continued until the budget ($B^*$) is exhausted. Therefore, barring the rough estimates at the first iteration, Algorithm~\ref{rolloutwbeliefa} cannot be parallelized for allocation. However, just like Algorithm~\ref{rolloutwbelief}, each processor can compute the recovery for a distinct initial condition ($s_0$) separately. Because of reduction in the parallelization in Algorithm~\ref{rolloutwbeliefa}, the solutions, even though high-quality, are computed at a slower rate. For our simulations, $B^*\leq9 \cdot 10^5$ and $\tilde \alpha\leq 10^5$ in Algorithm~\ref{rolloutwbeliefa}.

\begin{figure}
	\centering
	\includegraphics[width=\columnwidth]{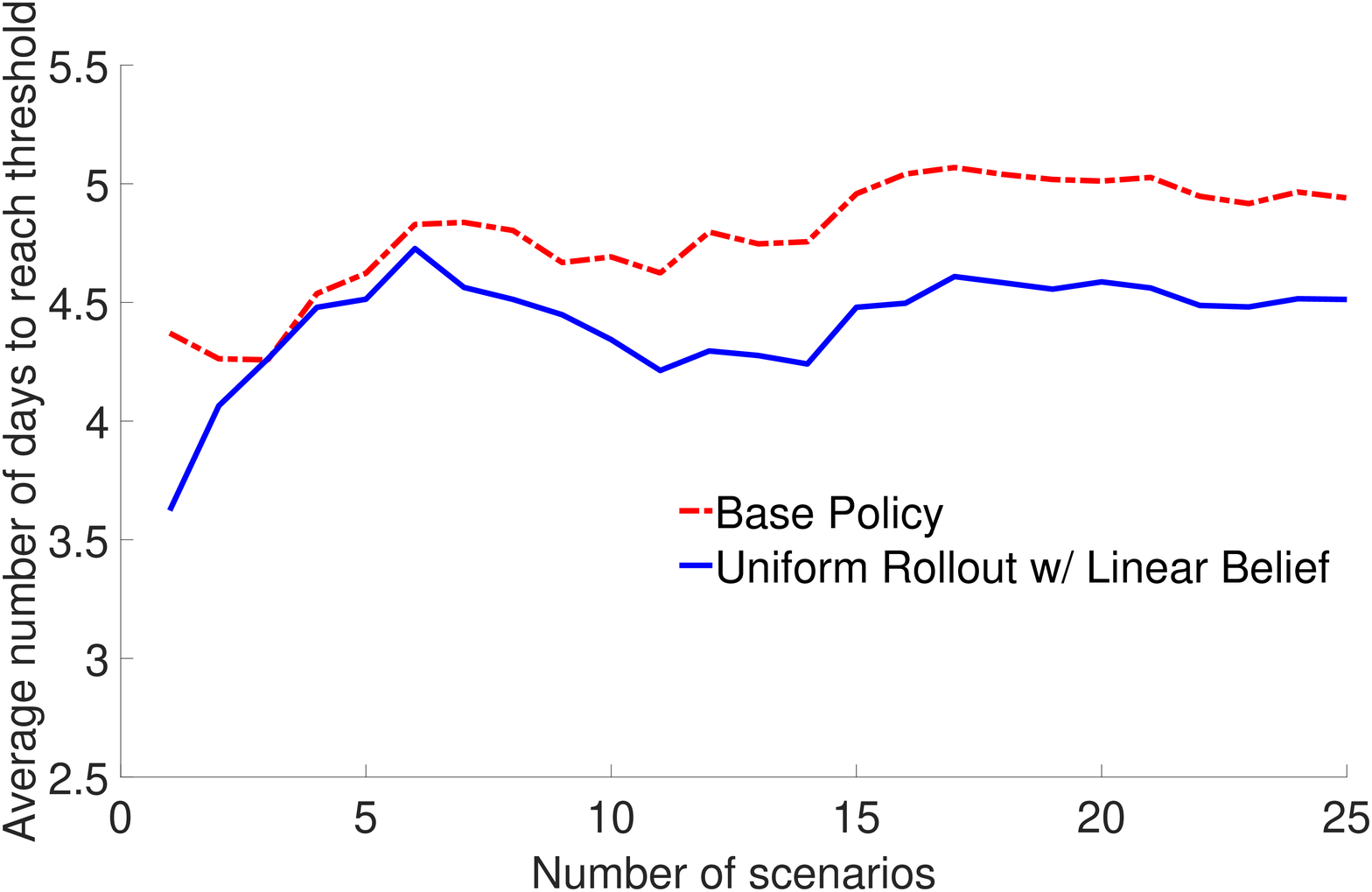}
	\caption{A cumulative moving average plot for the number of days required to provide electricity to 80\% of the population with respect to the total number of scenarios using Algorithm~\ref{rolloutwbelief}.}
	\label{fig3}
\end{figure}

Fig.~\ref{fig3} compares the performance of Algorithm~\ref{rolloutwbelief} with the base policy for Objective 1. For the simulations, $\zeta=0.8$; the goal is to calculate recovery actions so that 80\% of the population has electricity in minimum time. The figure depicts the cumulative moving average plot of the number of days required to achieve Objective 1. The cumulative moving average plot is computed by averaging the days required to reach the threshold for the total number of scenarios depicted on the X-axis of Fig.~\ref{fig3}. The cumulative moving average is used to smooth the data. As the number of scenarios increases in order to represent the stochastic behaviour of the earthquake model accurately, our algorithm saves about half a day over the recovery computed using the base policy. We manage to achieve the performance at scale (without any restriction on the number of workers, whereas all our earlier related work (see \cite{ress2,iEMSs,icasp}, \cite{ress1}, and \cite{case}) put a cap on the number of RUs); in addition, this performance is achieved on a local computational machine.

\begin{figure}	
	\centering
	\includegraphics[width=\columnwidth]{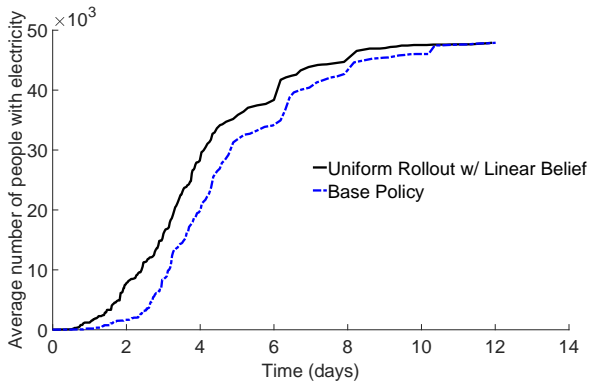}
	\caption{Average (of 25 recovery paths) recovery path using base policy and uniform rollout with linear belief for Objective 2.}
	\label{fig4}
\end{figure}

Fig.~\ref{fig4} compares the performance of Algorithm~\ref{rolloutwbelief} with the base policy for Objective 2. The recovery path (trajectories) for both the base policy and Algorithm~\ref{rolloutwbelief} are computed by calculating the average of 25 different recoveries over different initial conditions. The recovery path represents the number of people that have electricity after a given amount of time (days) because of recovery actions.  Evaluating the performance of our algorithm in meeting Objective 2 (defined in Section~\ref{probform}) boils down to calculating the area under the curve of our plots normalized by the total time for the recovery (12 days). The area represents the product of the number of people who have electricity after the completion of each repair action ($n_t$) and the time required in days for the completion of that action (the inter-completion time $r_t$). A larger value of this area ($\sum_{t}n_t\cdot r_t$) normalized by total time to recovery ($t_{\text{tot}}$) represents the situation where a greater number of people were benefitted as a result of the recovery actions. Normalization of the area ($\sum_{t}n_t\cdot r_t$) with the total time to recovery ($t_{\text{tot}}$) is important because the amount of time required to finish the recovery ($t_{\text{tot}}$) using the base policy and rollout with linear belief can be different. It is evident by visual inspection of the figure that recovery with Algorithm~\ref{rolloutwbelief} results in more benefit than its base counterpart; however, calculating $(\sum_{t}n_t\cdot r_t)/t_{\text{tot}}$ for the plots is necessary when the recovery achieved by the algorithms intersect at several points (see \cite{ress1}), a behaviour commonly seen with the rollout algorithm because of the lookahead property.

\begin{figure}
	\centering
	\includegraphics[width=\columnwidth]{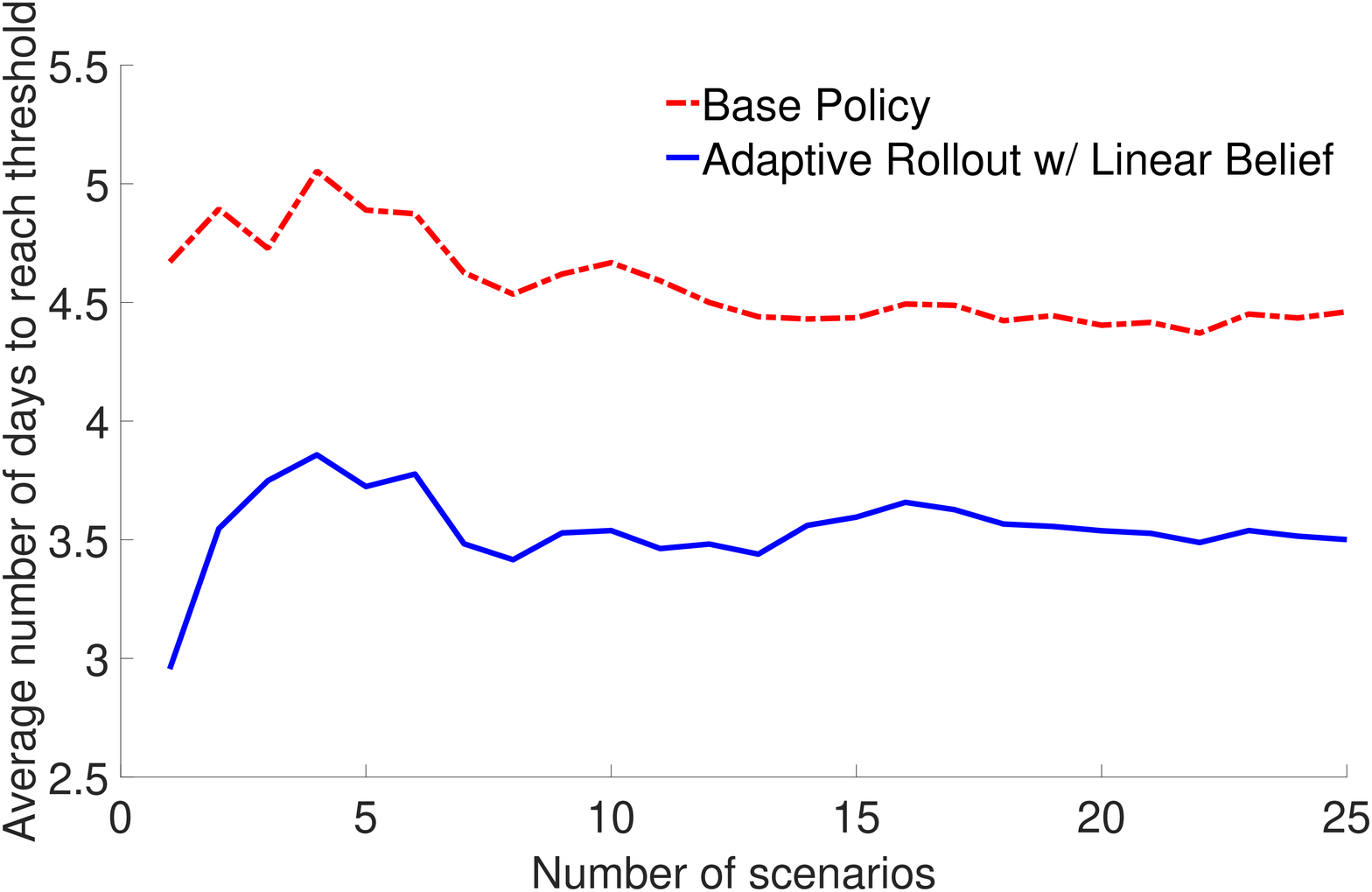}
	\caption{A cumulative moving average plot for the number of days required to provide electricity to 80\% of the population with respect to the total number of scenarios using Algorithm~\ref{rolloutwbeliefa}.}
	\label{fig5}
\end{figure}

Fig.~\ref{fig5} compares the performance of Algorithm~\ref{rolloutwbeliefa} with the base policy for Objective 1. Again, we set $\zeta=0.8$. In contrast to Algorithm~\ref{rolloutwbelief}, Algorithm~\ref{rolloutwbeliefa} improves the performance by another half a day so that the recovery because of its actions results in a saving of one day over the base policy to meet the objective. Adaptively allocating $B^*$ using UCB1, even though slower in runtime, can achieve better performance than Algorithm~\ref{rolloutwbelief} with a smaller simulation budget. In the end, the choice between Algorithm~\ref{rolloutwbeliefa} and Algorithm~\ref{rolloutwbelief} will be dictated by the urgency of the recovery action demanded from the automation framework and the computational platform deployed.

\begin{figure}
	\centering
	\includegraphics[width=\columnwidth]{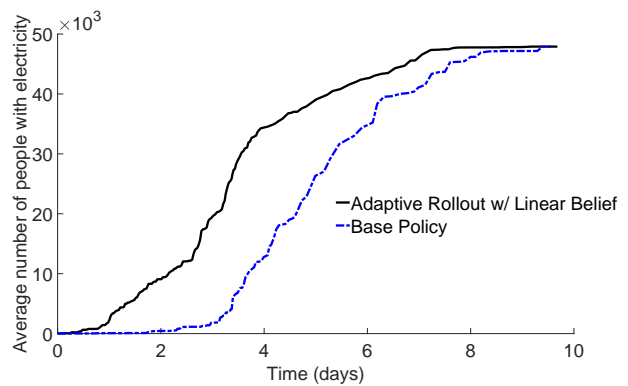}
	\caption{Performance comparison of adaptive rollout w/ linear belief vs. base policy for the second objective.}
	\label{fig6}
\end{figure}

Fig.~\ref{fig6} compares the performance of Algorithm~\ref{rolloutwbeliefa} with the base policy for Objective 2. Algorithm~\ref{rolloutwbeliefa} shows substantial improvement over the recovery calculated using both base policy and that using Algorithm~\ref{rolloutwbelief} in Fig.~\ref{fig4}. This is ascertained by calculating the area under the respective curves and normalizing it with the total time to recovery. Even though direct comparison between the recoveries of both the algorithms is not entirely appropriate owing to the stochastic initial conditions, random repair times, and a random base policy, it is worth re-noting that the performance of Algorithm~\ref{rolloutwbeliefa} is better than Algorithm~\ref{rolloutwbelief} at a lower simulation budget. Minimizing the cumulative regret to allocate $B^*$ during the parameter training provides for better recovery actions at each decision epoch. Because the entire framework is closed-loop, Algorithm~\ref{rolloutwbeliefa} (which uses both experiential and anticipatory learning) and Algorithm~\ref{rolloutwbelief} (which uses only anticipatory learning) exploit small improvements at each decision epoch $t$ and provides an enhanced recovery. Essentially, the small improvements squeezed at the earlier stages set a better platform for these algorithms to further exploit the anticipatory and experiential learning components at a later point in the recovery.

\section{Conclusion}
In this work, we presented a novel, systematic approach to MDPs that have jointly massive finite state and action spaces. When the action space consists of large number of discrete actions, the method of choice has been to embed these actions in continuous action spaces \cite{dulac2015deep}, where deep reinforcement learning techniques have shown promising performance on $|A|\approx10^6$. In contrast, in this study, we present a unique approach to address the problem, where the size of the discrete action space that we consider is significantly large than that in \cite{dulac2015deep}.

We studied an intricate real-world problem, modeled it in our framework, and demonstrated the powerful applicability of our algorithm on this challenging problem. The community recovery problem is a stochastic combinatorial decision-making problem, and the solution to such decision-making problems is critically tied with the welfare of communities in the face of ever-increasing natural and anthropogenic hazards. Our modeling of the problem is general enough to accommodate the uncertainty in the hazard models and the outcome of repair actions. Ultimately, we would like to test the techniques developed in this work on other real-world problems, e.g., large recommender systems (like those in use with the organizations YouTube and Amazon) and large industrial control systems.

\textbf{Ongoing Work:}
In our work on post-hazard community management (see \cite{ress2,iEMSs,icasp,emi}, \cite{ress1}, and \cite{case}), including this study, we have been focusing on obtaining solutions by the use of a single base policy. Currently, we are developing a framework where we leverage the availability of multiple base polices in the aftermath of hazards. Two algorithms are particularly appealing in this regard: parallel rollout and policy switching \cite{Chang}. In parallel rollout, just like in \cite{knowledge}, the optimization is done over the entire set $A(s_t)$. In our ongoing work, we are formulating a non-preemptive stochastic scheduling framework, where the size of set $A(s_t)$ grows linearly with the number of RUs, which circumvents the issue of large action spaces. In addition, we are also exploring heuristic search algorithms to guide the stochastic search, i.e., adaptively select the samples of the parallel rollout algorithm. There, we consider several infrastructure systems in a community, such as building structures, EPN, WN, and food retailers simultaneously (all these systems are inter-connected), and we compute the recovery of the community post-hazard.

\bibliography{IEEEabrv,IEEEexample}

\end{document}